# Evaluating Echo State Network for Parkinson's Disease Prediction using Voice Features

Seyedeh Zahra Seyedi Hosseininian [1], Ahmadreza Tajari [2], Mohsen Ghalehnoie [3*], Alireza Alfi [4]

[1] Faculty of Electrical Engineering, Shahrood University of Technology, Shahrood, Iran, zahra.seyyedihoseininian@gmail.com.

[2] Electrical Engineering Department, Sharif University of Technology, Tehran, Iran, ahmadreza.tajarri@ee.sharif.edu.

[3*] Faculty of Electrical Engineering, Shahrood University of Technology, Shahrood, Iran, ghalehnoie@shahroodut.ac.ir, +98-915-5448457 (corresponding author).

[4] Faculty of Electrical Engineering, Shahrood University of Technology, Shahrood, Iran, a_alfi@shahroodut.ac.ir.

**Abstract**

Parkinson's disease (PD) is a debilitating neurological disorder that necessitates precise and early diagnosis for effective patient care. This study aims to develop a diagnostic model capable of achieving both high accuracy and minimizing false negatives, a critical factor in clinical practice. Given the limited training data, a feature selection strategy utilizing ANOVA is employed to identify the most informative features. Subsequently, various machine learning methods, including Echo State Networks (ESN), Random Forest, k-nearest Neighbors, Support Vector Classifier, Extreme Gradient Boosting, and Decision Tree, are employed and thoroughly evaluated. The statistical analyses of the results highlight ESN's exceptional performance, showcasing not only superior accuracy but also the lowest false negative rate among all methods. Consistently, statistical data indicates that the ESN method consistently maintains a false negative rate of less than 8% in 83% of cases. ESN's capacity to strike a delicate balance between diagnostic precision and minimizing misclassifications positions it as an exemplary choice for PD diagnosis, especially in scenarios characterized by limited data. This research marks a significant step towards more efficient and reliable PD diagnosis, with potential implications for enhanced patient outcomes and healthcare dynamics.

1. Introduction

Parkinson's disease (PD) is a prevalent neurodegenerative disorder and the second most common after Alzheimer's disease, affecting approximately 1% of individuals over the age of 65 worldwide [1]. Initially identified by James Parkinson in 1817, PD is characterized by the degeneration of dopamine-producing neurons [2]. The insufficiency of dopamine leads to a variety of symptoms, including resting tremors, muscle stiffness, akinesia, bradykinesia (postural instability), and other significant features such as sleep disorders, cardiac arrhythmias, constipation, and changes in speech, which are crucial indicators of PD [3].

The accurate and timely diagnosis of Parkinson's disease is paramount for effective symptom management and appropriate treatment, as incorrect or delayed diagnosis can result in substantial physical and economic costs for patients [4]. Conversely, a precise diagnosis can prevent unnecessary interventions and reduce overall healthcare expenses. As a result, researchers and medical professionals in the field of Parkinson's disease diagnosis are actively seeking novel solutions that offer high diagnostic accuracy and optimal efficiency.

According to the authors' knowledge, current research on Parkinson's disease diagnosis can be broadly classified into three main categories. The first category involves diagnosing PD based on clinical symptoms and laboratory tests, such as single-photon emission computed tomography (SPECT) measurements and RNA-based methods [2], [5]. However, these medical tests can be expensive and not readily accessible. Furthermore, the analysis and interpretation of such tests often require expertise, introducing the possibility of human error in result interpretation.

The second category focuses on an expert examination of the physical symptoms displayed by affected individuals. However, the availability of experts may be limited, and human error can still be a factor [5]. Consequently, the third category involves leveraging artificial intelligence (AI) techniques to analyze symptoms and aid in diagnosis. Various approaches have been explored, including the application of the Relief technique to analyze three-dimensional SPECT images [6] and the utilization of convolutional neural networks for classifying electroencephalogram (EEG) images [7]. However, these methods often require expensive equipment for image recording and preparation.

Considering that tremor is one of the common symptoms of Parkinson's disease and can be recorded quickly and at a low cost, researchers have turned their attention towards utilizing it for diagnosis. Some studies have employed echo state networks (ESNs) to analyze tremors between two fingers [8] or employed the spiral test [9] to detect the presence of PD. Additionally, a study by [10] combined a decision tree with online Arabic handwriting to measure tremor levels and predict Parkinson's disease with 92.86% accuracy. However, although tremor can be a common symptom across various diseases, it cannot be solely relied upon for the diagnosis of Parkinson's disease [11]. Furthermore, the depletion of dopamine can cause voice disorders, which may manifest in the early stages of the disease before the appearance of tremors [12]. Research indicates that 89% of Parkinson's patients experience speech disorders, characterized by unclear expression, harsh and breathy voices, and monotonous pitch [13]. Consequently, recent AI-based research has focused on voice analysis as a means of diagnosing Parkinson's disease. For instance, Shamrat et al. utilized Support Vector Machines (SVM) to classify features extracted from voice recordings and predict Parkinson's disease [14]. In [13], employed AdaBoost in combination with an optimal data pre-processing approach to enhance prediction accuracy. Additionally, Imran Ahmed et al. used feature selection followed by the Random Forest algorithm to predict Parkinson's disease, thus reducing the complexity of the dataset [13]. Notably, despite the many capabilities of ESNs in speech recognition [15] and their applications in diagnosing various diseases such as breast cancer [16], predicting blood glucose concentration for type I diabetes [17], and detecting oral cancer [18], the use of ESNs, particularly in conjunction with feature reduction, has not been extensively explored in the context of voice analysis for Parkinson's disease diagnosis. This knowledge gap motivates our research to employ ESNs along with feature selection for diagnosing Parkinson's disease based on sound analysis.

Overall, this manuscript aims to investigate the application of an ESN neural network for the diagnosis of Parkinson's disease by using the features which are the least and the most significant. By extracting relevant features and leveraging the capabilities of ESNs, we aim to achieve an accurate and efficient diagnosis, contributing to the growing body of research in the field of AI-based approaches for Parkinson's disease diagnosis. So, the main contributions of this work are as follows:

1. Utilizing an echo state network for the diagnosis of Parkinson's disease. This approach represents a novel and effective method for diagnosing Parkinson's disease.

2. Extracting relevant features from voice recordings to improve the diagnostic accuracy of the ESN model. By employing feature selection techniques, we aim to identify the most informative features and reduce the complexity of the dataset, potentially enhancing the efficiency and accuracy of the diagnosis, especially when the number of training samples for learning is small.

3. Advancing AI-based approaches for Parkinson's disease diagnosis. By incorporating ESNs and feature selection techniques into the analysis of voice recordings, this research contributes to the growing field of AI-based approaches for Parkinson's disease diagnosis. This led to the development of more accurate, accessible, and cost-effective diagnostic tools, which improve patient outcomes and reduce healthcare costs.

The subsequent sections of the article are organized as follows. Section 2 presents the methodology employed, encompassing the various aspects of the study such as data preparation, the structure of the state echo network, feature extraction techniques, and the criteria used for evaluation. In Section 3, emphasis is placed on the implementation of feature extraction at multiple levels, the training process of the state echo network, and a comparative analysis of its performance against other established methodologies. The findings demonstrate the superior performance of the echo state network despite the limited number of training samples, thereby necessitating restricted input characteristics or features. Finally, Section 4 concludes the article by summarizing the key conclusions and implications derived from the study, accentuating the efficacy of the proposed approach, and suggesting potential avenues for further research and exploration.

## 2. Methodology
### 2.1. Data Preparation

This study involved a database of 31 individuals, 23 of whom were diagnosed with Parkinson's disease (PD) and sought treatment at the University of Oxford. The National Centre for Voice and Speech in Denver, Colorado recorded speech signals from these participants. Multiple types of voice samples, including sustained vowels, numbers, words, and short sentences, were collected from each subject, resulting in a total of 195 voice recordings (approximately six recordings per patient). From these recordings, a set of 22 linear and time-/frequency-based features were extracted, which were not independent. These features are detailed in Table 1.

The primary objective of this dataset is to develop a robust classification model capable of accurately differentiating between healthy individuals and PD patients. To achieve this, each individual in the dataset was assigned a one-bit label, where '0' represents a healthy individual and '1' represents a PD patient. Therefore, the dataset is suitable for classification purposes, as it provides a clear distinction between the two classes. Additional information about this dataset can be found in [19].

To mitigate the risk of overfitting in the classifier, the available data is randomly partitioned into two distinct sets: a training set consisting of 156 samples and a testing set comprising 39 samples. This partitioning approach allows the classifier to be trained on the training data and subsequently evaluated using the independent testing data. By separating the data into these sets, the classifier's performance can be effectively assessed without it becoming too specialized to the training data.

*Table 1. List of measurement methods applied to acoustic signals recorded from each subject of the used dataset* [19].

| Feature | Description |
|---|---|
| MDVP:Fo(Hz) | Average vocal fundamental frequency |
| MDVP:Fhi(Hz) | Maximum vocal fundamental frequency |
| MDVP:Flo(Hz) | Minimum vocal fundamental frequency |
| MDVP:Jitter(%) | Kay Pentax MDVP jitter as a percentage |
| MDVP:Jitter(Abs) | Kay Pentax MDVP absolute jitter in microseconds |
| MDVP:RAP | Kay Pentax MDVP Relative Amplitude Perturbation |
| MDVP:PPQ | Kay Pentax MDVP five-point Period Perturbation Quotient |
| Jitter:DDP | Average absolute difference of differences between cycles, divided by the average period |
| MDVP:Shimmer | Kay Pentax MDVP local shimmer |
| MDVP:Shimmer(dB) | Kay Pentax MDVP local shimmer in decibels |
| Shimmer:APQ3 | Three point Amplitude Perturbation Quotient |
| Shimmer:APQ5 | Five point Amplitude Perturbation Quotient |
| MDVP:APQ | Kay Pentax MDVP 11-point Amplitude Perturbation Quotient |
| Shimmer:DDA | Average absolute difference between consecutive differences between the amplitudes of consecutive periods |
| NHR | Noise-to-Harmonics Ratio |
| HNR | Harmonics-to-Noise Ratio |
| RPDE | Recurrence Period Density Entropy |
| DFA | Detrended Fluctuation Analysis |
| D2 | Correlation dimension |
| PPE | Pitch period entropy[20] |
| Spread1 | A type of nonlinear measure of fundamental frequency variation [20] |
| Spread2 | A type of nonlinear measure of fundamental frequency variation [20] |

It also helps prevent the classifier from memorizing the training samples and ensures a more generalizable model. However, it is important to note that as the number of input features in the classifier structure increases, both the computational load and the number of parameters requiring adjustment escalate. This makes it increasingly challenging to prevent overfitting and achieve proper parameter training. Therefore, a careful balance needs to be struck to select an appropriate number of input features to avoid overfitting and improve classifier performance.

Given the limited number of samples available in this study, including all 22 features as inputs in the classifier structure would be impractical. To overcome this challenge, a careful feature selection process is employed to identify a subset of the available features that are most informative for distinguishing between healthy individuals and PD patients. In this study, statistical analysis techniques such as ANOVA (Analysis of Variance) are utilized for feature selection. ANOVA is a statistical method used to determine the significance of differences between group means. In the context of this study, ANOVA is applied to evaluate the statistical significance of the relationships between each feature and the target variable (healthy or PD). Features that demonstrate a strong association with the target variable are selected for inclusion in the classifier.

The feature selection process is performed iteratively, gradually reducing the number of selected input features while maintaining their relevance to the classification task. This iterative approach helps to ensure that the selected subset of features captures the most salient characteristics of the speech signals related to PD diagnosis. To rigorously assess the performance of the proposed method and compare it with other established classifiers, several key evaluation metrics were employed. These metrics provide valuable insights into the classifier's ability to differentiate between healthy individuals and those with Parkinson's disease (PD), thus determining the diagnostic accuracy and reliability of the classifier. The chosen metrics include accuracy, precision, recall, and false negative rate.

Accuracy represents the ratio of correctly predicted instances to the total instances in the dataset. It provides a comprehensive overview of the overall classification performance, measuring both correct positive and negative predictions. Mathematically, accuracy is defined as follows:

$$Accuracy = \frac{True\ Positives\ (TP) + True\ Negatives (TN)}{Total\ Instances}. \tag{1}$$

Precision, also known as the positive predictive value, measures the accuracy of positive predictions made by the classifier. It quantifies the proportion of true positive predictions among all instances predicted as positive. Precision is particularly crucial in cases where false positives can have significant consequences. The formula for precision is given by:

$$Precision = \frac{True\ Positives\ (TP)}{True\ Positives\ (TP) + False\ Positives\ (FP)}. \tag{2}$$

Recall, also known as sensitivity or true positive rate, assesses the ability of the classifier to correctly identify all positive instances in the dataset. It calculates the proportion of true positive predictions among all actual positive instances. Recall is especially important when false negatives are critical to avoid. It can be calculated using the following formula:

$$Recall = \frac{True\ Positives\ (TP)}{True\ Positives\ (TP) + False\ Negatives\ (FN)}. \quad (3)$$

The false negative rate indicates the proportion of actual positive instances that were incorrectly predicted as negative by the classifier. This metric highlights the potential for missing important positive instances, which can be particularly significant in medical diagnosis. The false negative rate is calculated as:

$$False\ Negative\ Rate = \frac{False\ Negatives\ (FN)}{True\ Positives\ (TP) + False\ Negatives\ (FN)}. \quad (4)$$

By employing these metrics, we can gain valuable insights into the effectiveness and reliability of the proposed classifier [21].

Additionally, in the upcoming section, we delve into the detailed structure of the classifier employed in this study. Its architecture, algorithms, and underlying mechanisms will be elaborated on, highlighting the unique features that contribute to its performance. Furthermore, to provide a comprehensive comparison, the performance of the designed classifier will be contrasted with other commonly used classifiers, including Random Forest (RF), K-Nearest Neighbors (KNN), Support Vector Classifier (SVC), XGBoost (XGB), and Decision Tree (DT). These classifiers were selected due to their widespread usage in similar studies and their capability to handle classification tasks effectively. By comparing the performance of the proposed classifier with these established methods, we can assess its competitiveness and identify its strengths and weaknesses.

## 2.2. Structure of Echo-State-Network

In this subsection, we present the Echo State Network structure utilized as a classifier for Parkinson's disease (PD). ESN is a supervised learning algorithm designed for temporal data processing [22]. It is a type of recurrent neural network (RNN) introduced by Jager, specifically tailored for the prediction of non-linear time series and analysis of complex systems. The ESN consists of a three-layered network, including an input layer, a hidden layer known as the reservoir, and an output layer.

Figure 1 illustrates the schematic representation of the ESN structure. The input layer contains the input data, and the reservoir represents the hidden layer, consisting of sparsely connected internal units. The weights of the neurons in the reservoir are fixed and possess sparse random connectivity. The output layer, referred to as the readout layer, consists of neurons whose weights can be learned. This configuration allows the network to generate desired temporal patterns and adjust through learning processes.

One of the advantages of ESN is its compatibility with simple linear learning rules. During network initialization, the input and internal units' weights are randomly assigned and remain fixed. Consequently, during training, only the weights associated with the readout layer are adjusted. Despite its simple learning rule, the echo state network demonstrates the capability to tackle complex problems. By employing a sufficient number of internal units, the ESN expands the information from inputs into a higher-dimensional space, enabling optimal solution generation within a problem-solving domain. As a result, ESNs have found extensive application in various domains, including time series prediction, speech recognition, EEG signal analysis, brain modeling, and engineering fields [22].

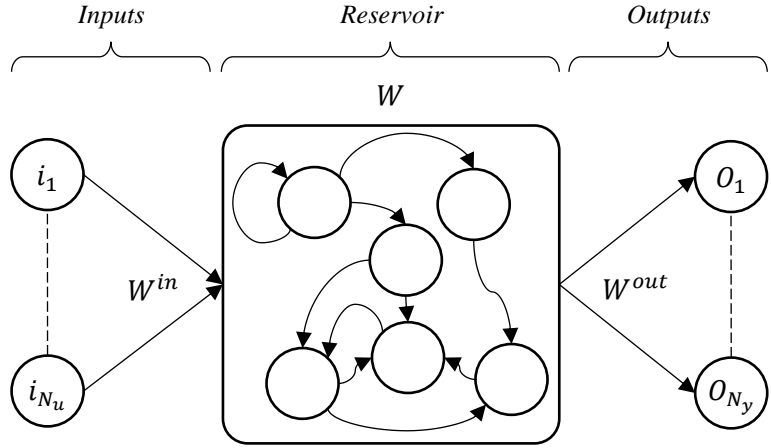

*Figure 1. Structure of a three-layered ESN [9].*

In the context of this study, the ESN structure is employed as a classifier for PD detection. The input neurons in the input layer receive data as a vector $u(n) \in \mathbb{R}^{N_u}$, where $N_u$ denotes the total number of input neurons. These inputs are connected to each reservoir neuron in the reservoir layer through the input matrix $W^{in} \in \mathbb{R}^{N_u \times N_x}$, where $N_x$ represents the total number of reservoir neurons. The activations of the reservoir neurons are stored in the vector $x$. The reservoir neurons are interconnected, and these connections are defined by the recurrent matrix $W \in \mathbb{R}^{N_x \times N_x}$. The output neurons are connected to both the reservoir and input neurons through the output matrix $W^{out} \in \mathbb{R}^{D \times N_y}$, where $D$ is the number of training data, and $N_y$ signifies the total number of output neurons. The output of the ESN is represented by the matrix $Y \in \mathbb{R}^{D*N_y}$ [23].

To implement an ESN and process an input sequence, several steps are considered. First, a large random reservoir is created, where the size of the reservoir $N_x$ is defined. It is important to avoid overfitting by ensuring that the number of reservoir neurons is not too large, i.e., $D < N_u + N_x$. The input and reservoir matrices are initialized with random values from a uniform or normal distribution.

Next, the activation of the reservoir neurons is calculated. For each time point n, the reservoir neurons' activations $\tilde{x}(n)$ and the reservoir activation $x(n)$ are updated using,

$$\tilde{x}(n) = f\left(W^{in}u(n) + Wx(n-1)\right). \tag{5}$$

The update is determined by the input matrix $W^{in}$, the current input $u(n)$, the recurrent weight matrix $W$, and the previous reservoir activation $x(n-1)$. The activation function $f(.)$, commonly chosen as the hyperbolic tangent ($tanh$), regulates the update process. Equation (6) defines the new reservoir activation, which incorporates the leaking rate $\alpha \in (0,1]$ and is and is a combination of the previous activation state and the reservoir update $\tilde{x}(n)$.

The leakage $\alpha$ is a control parameter for the leaky integration and adapts the input dynamics to the output dynamics.

$$x(n) = (1 - \alpha)x(n - 1) + \alpha\tilde{x}(n). \tag{6}$$

To calculate the output, the state matrix $X \in \mathbb{R}^{D \times (N_u + N_x)}$, which includes the inputs and reservoir activations, and the output matrix $W^{out}$ are utilized. The output is determined by multiplying the output matrix with the state matrix, as shown in,

$$Y = W^{out}X. \tag{7}$$

Typically, the state matrix $X$ is used after an initial time lag to mitigate initial effects. Unlike traditional neural networks, where all weights between layers are adjusted, ESN only requires training of the output weights $W^{out}$. The calculation of the output weights is based on ridge regression,

$$W^{out} = Y^{target}X^T(X^TX + \beta I) \tag{8}$$

where the target output $Y^{target}$, the state matrix $X$, the regularization coefficient $\beta$, and the identity matrix $I \in \mathbb{R}^{D \times D}$ are employed. The regularization coefficient $\beta$ is used to ensure invertibility [24]. The performance of ESN is influenced by several hyperparameters, including the reservoir size ($N_x$), spectral radius, leaking rate, and regression parameter. To achieve the best accuracy, we optimize these ESN hyperparameters using Bayesian optimization.

Lastly, as mentioned earlier, we compare the performance of ESN with other methods. These common classifiers are briefly explained below.

RF (Random Forest) is a machine-learning algorithm that utilizes an ensemble of decision trees. It differs from other methods by building multiple decision trees and randomly selecting subsets of features for each tree. The predictions of these trees are combined by aggregating their votes, and the class with the highest number of votes becomes the final prediction. RF is recognized for its ability to mitigate overfitting and capture nonlinear and interactive effects of variables. The parallelization of the training process significantly reduces the required training time. By combining the prediction results of each tree, RF effectively reduces variance and enhances overall prediction accuracy [25].

KNN (K-Nearest Neighbors) is a straightforward classification algorithm that operates based on the rule of nearest neighbors. It determines the class of test samples by calculating the distance between them and all the training samples, assigning the test samples to the class that is most prevalent among its k-nearest neighbors. KNN has been applied to analyze different voice features of PD patients, including jitter, shimmer, and harmonics-to-noise ratio, and has shown moderate accuracy rates in detecting Parkinson's disease [26].

SVC (Support Vector Classifier) is a conventional supervised machine learning technique widely used for classification and regression tasks. It relies on the kernel method and is characterized by its ability to split the decision boundary, effectively separating classes of data points. In the context of PD detection, SVC has demonstrated high accuracy rates by analyzing various movement features such as tremor amplitude, acceleration, and fast [27].

XGBoost (Extreme Gradient Boosting) is an ensemble learning algorithm that combines multiple machine learning models, specifically gradient-boosted decision trees. It builds multiple decision trees based on different decision criteria, similar to a random forest. XGBoost is known for its speed and efficiency, making it a preferred choice in various applications [8].

By comparing the performance of ESN with these common classifiers, we aim to evaluate its effectiveness in utilizing limited features for PD detection.

### 2.3. Feature Extraction

Utilizing feature selection techniques can aid in the visualization and comprehension of data, while also reducing the requirements for measurement and storage. Moreover, it can lead to shorter training and utilization times, thereby addressing the challenge of high-dimensional data, ultimately improving prediction accuracy [28], [29]. The application of feature selection algorithms becomes essential in eliminating irrelevant features within the feature space. This process of reducing features contributes to enhanced classification accuracy and reduces the execution time of the classifier [30].

In this research, we employ Analysis of Variance (ANOVA) to identify the $k$ most significant features, where $k$ represents the number of selected features. ANOVA is a statistical tool utilized to assess whether the mean values of multiple groups differ significantly from each other. It operates under the assumption of a linear relationship between variables and the target, as well as the normal distribution of these variables. To evaluate the equality of means statistically, the ANOVA technique employs F-tests. When it comes to feature selection, we can utilize the 'ANOVA F-value' obtained from this test to exclude specific features that are unrelated to the target variable [31].

The formula for the one-way ANOVA F-test statistic is,

$$F = \frac{explained\ variance}{unexplained\ variance} = \frac{between\_group\ variability}{within\_group\ variability}. \tag{9}$$

The "explained variance", or "between-group variability" is,

$$\sum_{i=1}^{K} \frac{n_i(\bar{Y}_i - \bar{Y})^2}{k-1}, \tag{10}$$

where $\bar{Y}_i$ denotes the sample mean in the i-th group, $n_i$ is the number of observations in the i-th group, $Y_i$ denotes the overall mean of the data, and K denotes the number of groups. The "unexplained variance", or "within-group variability" is,

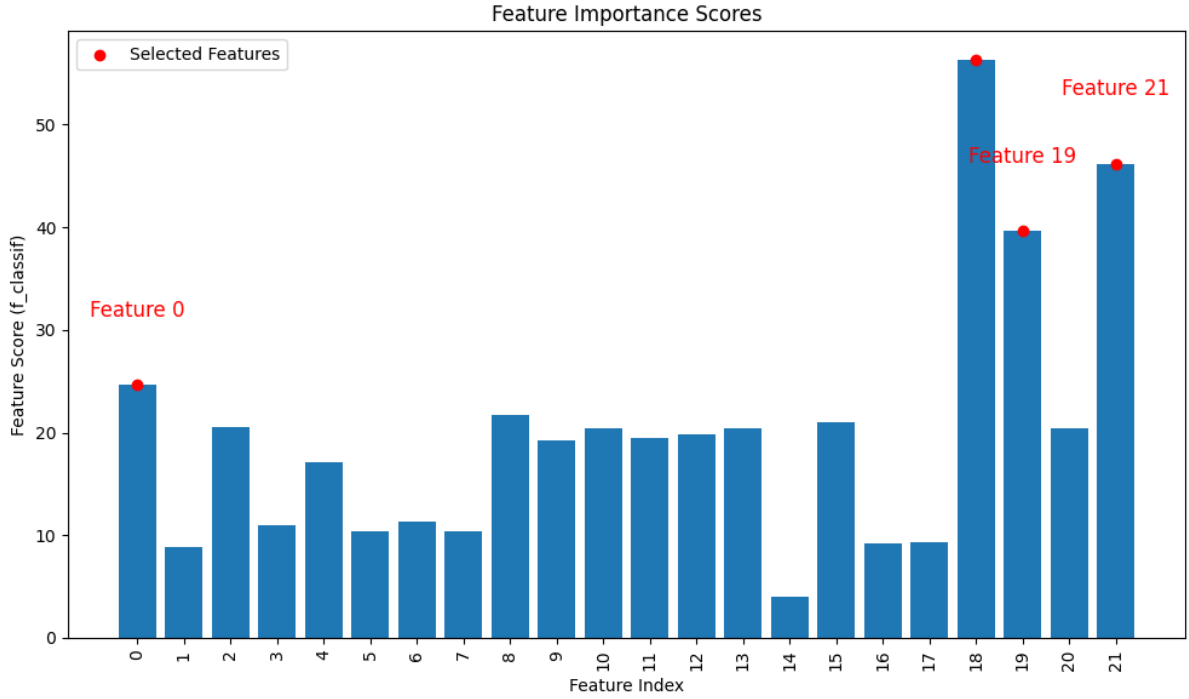

*Figure 2. Feature significance analysis using ANOVA F-Test where higher F-values denote greater feature significance.*

$$\sum_{i=1}^{K} \sum_{j=1}^{n_i} \frac{(Y_{ij} - \overline{Y}_i)^2}{N - K} \tag{11}$$

where $Y_{ij}$ is the $j^{th}$ observation in the $i^{th}$ out of $K$ groups and $N$ is the overall sample size [31].

Figure 2 present the F-values corresponding to each feature. This visualization highlights the features with the highest F-classic scores, emphasizing their significance in our feature selection process. The meticulous implementation of this method facilitated the reduction of our initial set of 22 features to a compact and pivotal subset comprising only four variables: MDVP:Fo(Hz), spread1, spread2, and PPE. By using ANOVA to carefully choose features and combining it with powerful ESNs, we are making a big leap forward in AI-based Parkinson's disease diagnosis.

3. **Simulation Results and Discussion**

In this section, we present the simulation results obtained from a comprehensive evaluation of six distinct methods: ESN, RF, KNN, SVC, XGB, and DT. In order to evaluate each of the methods, first, we selected a 20% random subset from the database, which was then utilized to evaluate the performance of each method, as test dataset .the trainable parameters of each of these methods are adjusted through Bayesian optimization. To ensure a comprehensive assessment, we initially fine-tuned the trainable parameters of each method through Bayesian optimization, which are the remaining 80% data from the available database as a trainig test. This evaluation process was meticulously repeated for 100 trials, and the resulting mean and standard deviation for each evaluation metric were meticulously calculated.

Table 2. The mean and standard deviation of evaluation metrics across 100 trials.

| method | Accuracy (%) mean | Std. Dev. | Precision (%) mean | Std. Dev. | Recall (%) mean | Std. Dev. | False Negative (%) mean | Std. Dev. |
|---|---|---|---|---|---|---|---|---|
| ESN | 88.185 | 3.607 | 90.363 | 4.208 | 94.489 | 2.794 | 5.511 | 2.794 |
| RF | 87.735 | 4.117 | 90.205 | 4.104 | 94.027 | 3.492 | 5.973 | 3.492 |
| KNN | 89.065 | 3.602 | 92.545 | 3.610 | 93.034 | 3.927 | 6.966 | 3.927 |
| SVC | 84.855 | 3.403 | 86.238 | 4.027 | 95.210 | 3.680 | 4.790 | 3.680 |
| XGB | 88.375 | 4.274 | 90.820 | 4.056 | 94.140 | 3.670 | 5.860 | 3.670 |
| DT | 84.094 | 4.808 | 90.891 | 4.258 | 87.933 | 5.535 | 12.067 | 5.535 |

Table 3. Evaluation of the various studied methods using F1-Score metric.

| | ESN | RF | KNN | SVC | XGB | DT |
|---|---|---|---|---|---|---|
| F1-Score | 92.380 | 92.076 | 92.790 | 90.502 | 92.450 | 89.388 |

The evaluation metrics utilized in this analysis include accuracy, positive precision, sensitivity (recall), and false negative rate. These metrics provide a comprehensive view of the methods' performance in diagnosing Parkinson's disease based on the selected acoustic features. Detailed results are thoughtfully summarized in Table 2.

Upon an initial examination of the Accuracy metric in Table 2, it appears that the KNN method exhibits superior performance. However, it's noteworthy that the ESN, RF, and XGB methods also demonstrate admirable performance in diagnosing Parkinson's disease using the four selected features, closely rivaling the KNN method. Moreover, both the KNN and ESN methods demonstrate commendable robustness, as indicated by their minimal standard deviations in accuracy. Nevertheless, it's imperative to acknowledge that the existing Parkinson's disease database and its statistical representation in various human societies are inherently imbalanced. Consequently, relying solely on the Accuracy metric may not yield a comprehensive interpretation of method superiority. In such scenarios, careful consideration of Precision and Recall metrics is crucial.

Upon referencing Table 3, which presents the F1-Score criterion (a harmonic mean of Precision and Recall) for all examined methods, it becomes apparent that there are no substantial differences among the top four methods under investigation. Hence, to gather a more nuanced interpretation, we must explore additional statistical criteria.

In the following, we obtain the cumulative density function (cdf) pertaining to the accuracy of each method. These cumulative distributions, coupled with a probability density function estimate of accuracy, are illustrated in Figure 3. Additionally, quantitative insights regarding the cumulative density function of accuracy are thoughtfully summarized in Table 4, providing a more comprehensive perspective on the performance of these methods in diagnosing Parkinson's disease.

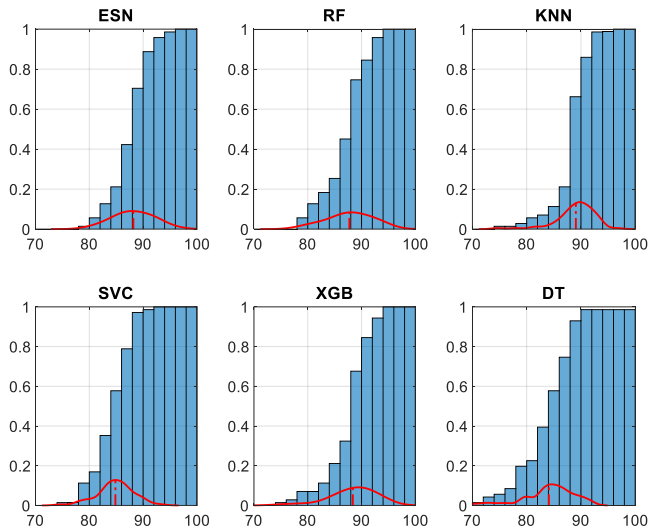

*Figure 3. The histogram of cumulative density function for accuracy and the probability density function estimate in different methods.*

Upon the probability density function estimate and its cumulative distribution pertaining to the studied methods (refer to Figure 3 and Table 4), among the top four methods, it is evident that only two methods, ESN and RF, consistently achieve an accuracy rate exceeding 78%. In contrast, the KNN method exhibits accuracy below 78% in a mere 1.5% of the evaluation trials, while the XGB method records accuracy below this threshold in approximately 2.8% of cases. Conversely, RF and XGB methods never surpass an accuracy of 96%, whereas ESN and KNN have achieved superior accuracy in approximately 1.5% of cases.

*Table 4. The accuracy's cumulative density for the studied methods is obtained from 100 trials.*

| method | 70%-72% | 72%-74% | 74%-76% | 76%-78% | 78%-80% | 80%-82% | 82%-84% |
|---|---|---|---|---|---|---|---|
| ESN | 0 | 0 | 0 | 0 | 0.0141 | 0.0563 | 0.1268 |
| RF | 0 | 0 | 0 | 0 | 0.0563 | 0.1268 | 0.1831 |
| KNN | 0 | 0 | 0.0141 | 0.0141 | 0.0282 | 0.0563 | 0.0704 |
| SVC | 0 | 0 | 0.0141 | 0.0141 | 0.1127 | 0.169 | 0.3521 |
| XGB | 0 | 0 | 0.0141 | 0.0282 | 0.0704 | 0.0704 | 0.1127 |
| DT | 0.0141 | 0.0423 | 0.0563 | 0.0845 | 0.1972 | 0.2254 | 0.3944 |
| method | 84%-86% | 86%-88% | 88%-90% | 90%-92% | 92%-94% | 94%-96% | 96%-98% |
| ESN | 0.2113 | 0.4225 | 0.7042 | 0.8873 | 0.9577 | 0.9859 | 1 |
| RF | 0.2535 | 0.4507 | 0.7465 | 0.8451 | 0.9577 | 1 | 1 |
| KNN | 0.1127 | 0.2113 | 0.662 | 0.8592 | 0.9859 | 0.9859 | 1 |
| SVC | 0.5775 | 0.7887 | 0.9718 | 0.9859 | 1 | 1 | 1 |
| XGB | 0.2113 | 0.3239 | 0.6761 | 0.8451 | 0.9437 | 1 | 1 |
| DT | 0.5775 | 0.9296 | 0.9859 | 0.9859 | 0.9859 | 0.9859 | 0.9859 |

So, in the context of upper and lower bounds for accuracy, the ESN method generally demonstrates superiority. However, it is essential to acknowledge the infrequent occurrence of these upper and lower bounds prevents a definitive conclusion regarding the ESN's superiority. For instance, in ESN, accuracy falls below 90% in approximately 70% of cases, resulting in higher accuracy in only 30% of instances. Conversely, in the case of the XGB and KNN methods, this phenomenon transpires in roughly 33% of cases, offering a slight edge in this regard. The variability in method superiority across different facets of the accuracy metric underscores the database's inherent imbalance and highlights the necessity for comprehensive statistical analyses across other metrics.

It's imperative to recognize that when two different methods yield almost identical levels of accuracy, this suggests that the sum of True Positives (TP) and True Negatives (TN) is comparable in both methods, although the distribution between TN and TP may differ. A higher TP count corresponds to a lower False Negative (FN) count, while an elevated TN count corresponds to a reduced False Positive (FP) count. In the context of disease diagnosis, misclassifying a healthy individual as sick imposes financial costs for further diagnostic procedures, but the opposite potentially leading to irreversible health consequences. Thus, the Recall metric, specifically the False Negative Rate, holds significant importance in distinguishing between methods with similar accuracy levels.

Among the four highest-accuracy diagnostic methods, as delineated in Table 2, the ESN method outperforms others in terms of the False Negative Rate. This advantage is further underscored by the probability density function and cumulative distribution information, as depicted in Figure 4 and Table 5. For instance, ESN is the sole method with a false detection rate below 14% in 100% of runs, and this error rate falls below 8% in 83% of runs. Consequently, in diagnosing Parkinson's disease based on the four audio characteristics examined in this research, it is evident that the ESN method excels in achieving a harmonious balance between diagnostic accuracy and the ability to correctly identify both healthy and afflicted individuals.

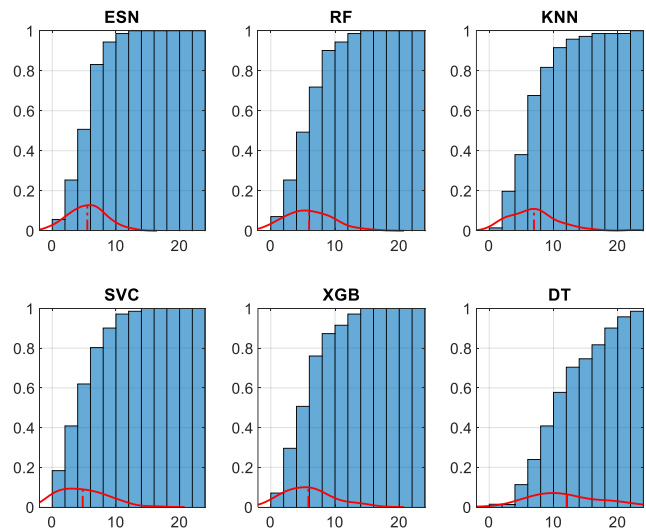

*Figure 4. The histogram of cumulative density function for False Negative Rate and the probability density function estimate in different methods.*

Table 5. The FN rate's cumulative density for the studied methods is obtained from 100 trials.

| method | 0%-2% | 2%-4% | 4%-6% | 6%-8% | 8%-10% | 10%-12% |
|---|---|---|---|---|---|---|
| ESN | 0.0563 | 0.2535 | 0.507 | 0.831 | 0.9437 | 0.9859 |
| RF | 0.0704 | 0.2535 | 0.493 | 0.7183 | 0.9014 | 0.9437 |
| KNN | 0.0141 | 0.1972 | 0.3803 | 0.6761 | 0.8169 | 0.9155 |
| SVC | 0.1831 | 0.4085 | 0.6197 | 0.8028 | 0.9014 | 0.9718 |
| XGB | 0.0704 | 0.2958 | 0.507 | 0.7606 | 0.8732 | 0.9155 |
| DT | 0.0141 | 0.0141 | 0.1127 | 0.2394 | 0.4085 | 0.5775 |
| method | 12%-14% | 14%-16% | 16%-18% | 18%-20% | 20%-22% | 22%-24% |
| ESN | 1 | 1 | 1 | 1 | 1 | 1 |
| RF | 0.9859 | 1 | 1 | 1 | 1 | 1 |
| KNN | 0.9577 | 0.9718 | 0.9859 | 0.9859 | 0.9859 | 1 |
| SVC | 0.9859 | 1 | 1 | 1 | 1 | 1 |
| XGB | 0.9718 | 1 | 1 | 1 | 1 | 1 |
| DT | 0.7042 | 0.7465 | 0.8169 | 0.9014 | 0.9577 | 0.9859 |

## 4. Conclusion

This study represents a significant advancement in the field of Parkinson's disease (PD) diagnosis through the utilization of machine learning. Our exploration of various algorithms unequivocally establishes the ESN model as the frontrunner, with a particular emphasis on both accuracy and minimizing false negative rates. The synergy achieved between ESN and feature extraction underscores a groundbreaking stride in AI-driven PD diagnosis, especially in scenarios with limited training data and economic constraints. This analytical framework seamlessly aligns with the evolving landscape of medical AI, poised to revolutionize practical and resource-efficient diagnostic protocols. Looking ahead, our chosen path holds promise in refining diagnostic precision and establishing a robust foundation for early PD detection.

**Conflict of interest statement**

The authors have no competing interests to declare that are relevant to the content of this article.